\pdfoutput=1

\documentclass[11pt]{article}

\usepackage[preprint]{acl}

\usepackage{times}
\usepackage{latexsym}

\usepackage[T1]{fontenc}

\usepackage[utf8]{inputenc}

\usepackage{microtype}

\usepackage{inconsolata}

\usepackage{graphicx}

\usepackage{adjustbox}
\usepackage{makecell}
\usepackage{multicol}
\usepackage{multirow}
\usepackage{hhline}
\usepackage{booktabs}
\usepackage{color}
\usepackage{xcolor}
\usepackage{amssymb}
\usepackage{amsmath}
\usepackage{CJK}

\newcolumntype{L}[1]{>{\raggedright\arraybackslash}p{#1}}
\newcolumntype{C}[1]{>{\centering\arraybackslash}p{#1}}
\newcolumntype{R}[1]{>{\raggedleft\arraybackslash}p{#1}}

\DeclareSymbolFont{extraup}{U}{zavm}{m}{n}
\DeclareMathSymbol{\varheart}{\mathalpha}{extraup}{86}
\DeclareMathSymbol{\vardiamond}{\mathalpha}{extraup}{87}
\DeclareMathSymbol{\varclubsuit}{\mathalpha}{extraup}{88}

\definecolor{purple}{RGB}{128, 0, 128}

\title{Detoxification of Large Language Models through\\Output-layer Fusion with a Calibration Model}

\author{%
    Yuanhe Tian$^{\varheart *}$, \hspace{0.1cm}
    Mingjie Deng$^{{\spadesuit}*}$, \hspace{0.1cm}
    Guoqing Jin$^{\Diamond}$, \hspace{0.1cm}
    Yan Song$^{{\spadesuit}\dag}$
    \\
    $^{\varheart}$University of Washington 
    \hspace{0.1cm}
    $^{\spadesuit}$University of Science and Technology of China \\
    $^{\Diamond}$People’s Daily Online
    \\
    $^{\varheart}$\texttt{yhtian@uw.edu} \hspace{0.1cm}
    $^{\spadesuit}$\texttt{dmj123456@mail.ustc.edu.cn} \\
    $^{\Diamond}$\texttt{jinguoqing@people.cn} \hspace{0.1cm}
    $^{\spadesuit}$\texttt{clksong@gmail.com} 
}

\begin{document}
\maketitle

\renewcommand{\thefootnote}{\fnsymbol{footnote}}
\footnotetext[1]{Equal contribution.}
\footnotetext[2]{Corresponding author.}

\renewcommand{\thefootnote}{\arabic{footnote}}

\begin{abstract}

Existing approaches for Large language model (LLM) detoxification generally rely on training on large-scale non-toxic or human-annotated preference data, designing prompts to instruct the LLM to generate safe content, or modifying the model parameters to remove toxic information, which are computationally expensive, lack robustness, and often compromise LLMs' fluency and contextual understanding.
In this paper, we propose a simple yet effective approach for LLM detoxification, which 
leverages a compact, pre-trained calibration model that guides the detoxification process of a target LLM via a lightweight intervention in its generation pipeline.
By learning a detoxified embedding space from non-toxic data, the calibration model effectively steers the LLM away from generating harmful content.
This approach only requires a one-time training of the calibration model that is able to be seamlessly applied to multiple LLMs without compromising fluency or contextual understanding.
Experiment results on the benchmark dataset demonstrate that our approach reduces toxicity while maintaining reasonable content expression.\footnote{The code and related resources about this paper are released at \url{https://github.com/synlp/LLM-Detoxic}.}
\newline
\textbf{Disclaimer}: This paper contains examples of hateful speech, which are used solely for evaluating and demonstrating the proposed model and do not reflect the authors’ personal views or endorsement of such language.
\end{abstract}

\section{Introduction}

Pre-trained language models have achieved impressive milestones in natural language generation (NLP) tasks \cite{devlin-etal-2019-bert,tian-etal-2020-supertagging,song2021zen,taori2023alpaca,touvron2023llama-2,achiam2023gpt,gan2023ziya2,wu2024taiyi}. 
However, limited controllability often leads to undesirable outputs, with toxicity being a significant concern. 
Such harmful outputs not only risk spreading misinformation and unethical content but also undermine user trust and AI acceptance. 
Mitigating toxicity, especially for large language models (LLMs), is thus crucial for ethical AI use and quality human–computer interaction \cite{geva2022transformer,liu2023context,tian2023chimed,huang2024chat,youssef2025position}.

\begin{figure*}
    \centering
    \includegraphics[width=1\linewidth]{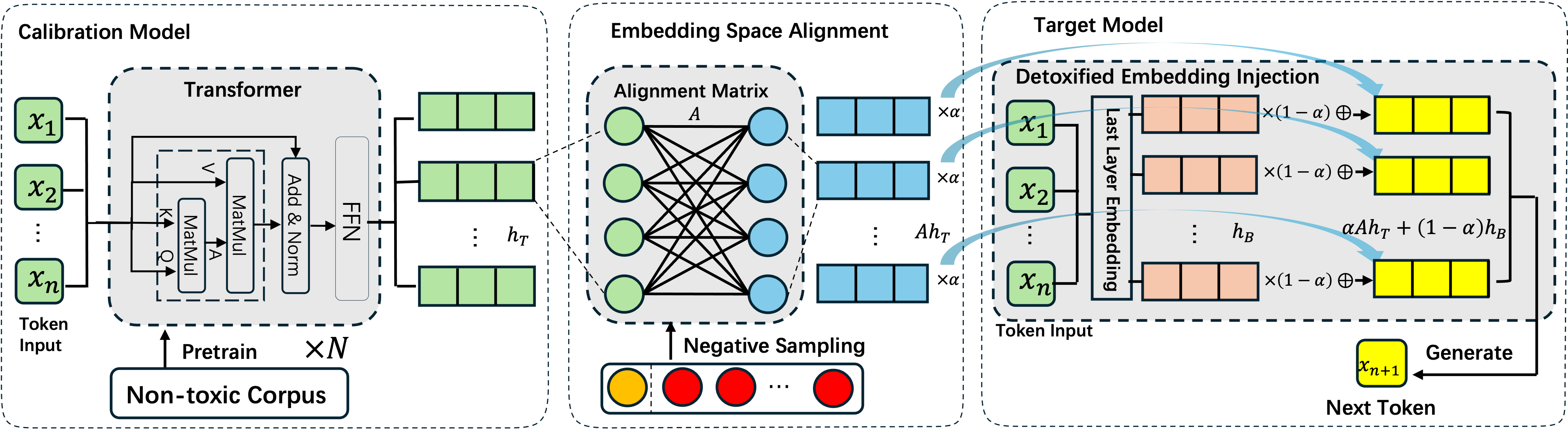}
    \caption{The overall architecture of our approach. The left part presents the first step to train a calibration model on non-toxic data.
    The middle part illustrates the second step aligns the embedding space of the calibration model with the target model.
    The right part shows the third step to inject the aligned detoxified embedding into the original LLM to reduce the toxicity in the generated content.}
    \label{fig:model}
\end{figure*}

In general, to rectify and ensure the safety of LLMs' generation outputs,
the straightforward strategy is to train them on specific data \cite{gehman2020realtoxicity,geva2022transformer,wang2022exploring,lu2022quark,korbak2023pretraining} using a pre-training and fine-tuning paradigm, or utilize reinforcement learning \cite{ouyang2022training,rafailov2023dpo} to align LLMs with human preference, which demands massive computation requirement \cite{wang2022exploring,lu2022quark}. 
To reduce such costs, alternatives that avoid full retraining have been proposed \cite{dathathri2020plug,10.1162/tacl_a_00434,meng2022locating,leong2023self,panickssery2023steering,niu2024parameter,qiu2024spectral,wang2024detoxifying,li2024precision,kim2024decoupling},
where one research line uses prompt-based detoxification \cite{dathathri2020plug,krause2020gedi,leong2023self} to guide generation at runtime without altering model parameters, though this requires heavy prompt engineering for different LLMs, 
and the other directly edits model parameters or the hidden vectors when running the model to remove toxic subspaces \cite{uppaal2024detox,wang2024detoxifying,li2024precision,kim2024decoupling}, which permanently changes the model and risks unintended side effects, such as altering its underlying knowledge.
Both lines have their limitation in either unstable prompting or complex model editing operations.
Therefore, it requires more lightweight and easy-to-control approaches for LLM detoxification.

In this paper, we propose a simple detoxification adaptation that directly refines the internal representations of an LLM to mitigate toxicity.
Our approach leverages a small calibration model, pre-trained on a curated non-toxic corpus to learn an embedding space so that provides contextualized representations for detoxification guidance.
Then we propose a fast embedding alignment method based on negative sampling to bridge the gap between the representations from the calibration model and those of the target LLM.
As a result, the aligned detoxified embeddings are injected into the last layer of the target LLM to
shape its generation process toward non-toxic outputs.
In doing so, our approach not only changes target LLMs' latent attributes in a direct manner but also avoids the computational overhead of re-training LLMs or the complexity of intricate parameter edits.
Experiment results on the benchmark dataset demonstrate the validity of our approach, which is able to reduce the toxicity without hurting the content of the generated output.

\section{The Approach}

Our approach mitigates the toxicity of a target LLM by refining its internal representations through a straightforward process, where a compact calibration model and an embedding alignment process are utilized.
The overall workflow is illustrated in Figure \ref{fig:model}, \textcolor{black}{where there are three main steps in our approach.
As the first step (which is illustrated on the left of Figure \ref{fig:model}), the compact model is pre-trained on a curated, safe-text corpus to learn a detoxified embedding space.
Then, we perform embedding alignment (which is presented on the middle of Figure \ref{fig:model}) with a negative sampling strategy to bridge the semantic space between the detoxified embeddings produced by the calibration model and those of the target LLM.
Finally, the aligned detoxified embedding is injected into the target LLM to shape its generation process toward non-toxic outputs, where the process is demonstrated on the right of Figure \ref{fig:model}.
}
The following subsections detail each component of our approach.

\subsection{Detoxified Embedding Pre-training}

As the first step, a compact calibration model is employed to learn a detoxified embedding space from non-toxic data.
Specifically, given an input text sequence $\mathbf{x} = (x_1, x_2, \ldots, x_N)$ with segmentation into discrete tokens \textcolor{black}{(the $n$-th token is denoted as $x_n (1\leq n\leq N)$)},
we map the tokens to continuous vectors using an embedding matrix $\mathbf{E} \in \mathbb{R}^{V \times d}$, where $V$ is the vocabulary size and $d$ is the embedding dimension.
Consider pre-trained embeddings generally learns the correlation between tokens \cite{pennington-etal-2014-glove,song-etal-2017-learning,song2018joint,song2018complementary,peters-etal-2018-deep,han-etal-2018-hyperdoc2vec,song-etal-2018-directional} (e.g., the Word2Vec \cite{mikolov2013efficient} approach learns the co-occurrence of tokens in a particular context window), we utilize pre-trained token embeddings instead of randomly initialized ones.
This design allows the calibration model to focus more on removing toxicity rather than learning language modeling.
These token embeddings, denoted as $\mathbf{e}_1, \mathbf{e}_2, \ldots, \mathbf{e}_N$, serve as the initial representation of the input.
The embeddings are subsequently processed by a Transformer with $L$ layers to generate the output $\widehat{\mathcal{Y}}$ through the standard auto-regressive process, where self-attention mechanisms and feed-forward networks refine the representations at each layer. 
The output $\widehat{\mathcal{Y}}$ is compared with the desired non-toxic output $\mathcal{Y}^{*}$ to compute the cross-entropy loss $\mathcal{L}$, \textcolor{black}{which is used to update the model parameters through an optimization approach (e.g., the Adam \cite{kingma2014adam}).}
Therefore, the training favors a detoxified embedding space that naturally downweights the toxic features.
The resulting detoxified embedding (denoted as $\mathbf{E}^{de}$) is used as guidance to align and inject non-toxic representations into the target LLM.

\subsection{Embedding Space Alignment}

A challenge arises when the embedding space of the calibration model and the target LLM differ from each other.
Therefore, we utilize an embedding space alignment solution to this challenge with negative sampling.
In doing so, we identify a set of tokens that are common to the tokenizers of both models.
For each token, we designate it as a positive example and randomly select a set of negative tokens (denoted as $\{t^-_1, t^-_2, \dots, t^-_K\}$) from the common vocabulary. 
The detoxified calibration model utilize the detoxified embedding $\mathbf{E}^{de}$ to produce an embedding for the positive token, $\mathbf{e}^+$, which is then transformed by an alignment matrix $\mathbf{A}$ to yield an aligned embedding vector
$\mathbf{z} = \mathbf{A} \cdot \mathbf{e}^+$.
Concurrently, the target LLM provides its own embedding for the positive token, $\mathbf{e}^{+*}$, as well as embeddings for the negative tokens, $\mathbf{e}^{-*}_{1}, \mathbf{e}^{-*}_{2}, \dots, \mathbf{e}^{-*}_{K}$. 
We compute the inner products between $\mathbf{z}$ and these target embeddings and formulate an objective that maximizes the similarity for the positive example while minimizing it for the negative ones,
resulting in the alignment loss
\begin{equation}
    \mathcal{L}_{\text{align}} = -\log \frac{\exp\bigl(\mathbf{z} \cdot \mathbf{e}^{+*}\bigr)}{\exp\bigl(\mathbf{z} \cdot \mathbf{e}^{+*}\bigr) + \sum_{k=1}^{K} \exp\bigl(\mathbf{z} \cdot \mathbf{e}^{-*}_{k}\bigr)}
\end{equation}
and we train the alignment matrix $\mathbf{A}$ by minimizing this loss, so as to ensure that the detoxified embedding from the calibration model is properly mapped to the target LLM’s embedding space.

\subsection{Detoxified Embedding Injection}

LLMs achieve outstanding performance on many NLP tasks \cite{taori2023alpaca,touvron2023llama-2,achiam2023gpt,tian-etal-2024-dialogue-summarization,tian-etal-2024-learning-multimodal,li-etal-2024-challenging}.
However, they may generate toxic information, as LLMs assign high probabilities to harmful tokens in their output layer.
Upon the completion of training and alignment, we inject the resulting detoxified embedding into target LLMs to influence the generation process.
For a given prompt $\mathbf{x}$, both the aligned calibration model and the base LLM independently process the input to produce their final-layer hidden representations, denoted by $\mathbf{h}_T$ and $\mathbf{h}_B$, respectively.
Then, we compute an aggregated embedding via a weighted combination, which is formulated as
\begin{equation} \label{eq:agg}
    \mathbf{h}_{\text{agg}} = \alpha\, \mathbf{A} \cdot \mathbf{h}_T + (1 - \alpha)\, \mathbf{h}_B
\end{equation}
where $\alpha \in [0,1]$ is a hyper-parameter that balances the contribution of the detoxified embedding and the target model’s original output representation.
As a result, the aggregated embedding $\mathbf{h}_{\text{agg}}$ replaces the original final-layer hidden state of the target model prior to the decoding phase,
where such fusion ensures that the subsequent generation process is conditioned on a representation that is both contextually rich and non-toxic. 

\section{Experiments}

\subsection{Settings}

In this work, we employ two datasets for pre-training the detoxified embeddings and evaluating the toxicity reduction performance of our approach, respectively.
The first dataset, WildJailbreak\footnote{We get the data from \url{https://huggingface.co/datasets/allenai/wildjailbreak}.} \cite{jiang2025wildteaming}, is a large-scale safety training corpus released by the AI$^2$ laboratory.
It contains approximately 262,000 training examples that are carefully curated to cover a wide range of safety-critical scenarios and are designed to promote the learning of non-toxic and responsible language patterns.
The second dataset is RealToxicityPrompts\footnote{We get the data from \url{https://huggingface.co/datasets/allenai/real-toxicity-prompts}.} \cite{gehman2020realtoxicity}, which is widely recognized in the community for its challenging set of inputs that are known to trigger toxic generations in language models.
For evaluating the toxicity in LLMs, we utilize a subset of the RealToxicityPrompts dataset, namely the \texttt{challenge\_prompts} collection, which contains 1,199 prompts that are most likely to expose the weaknesses of a model with respect to toxic output.

\begin{table}[t]
\centering
\begin{tabular}{l|cc}
\toprule
\textbf{Model}
  & \textbf{Toxicity} ($\downarrow$)
  & \textbf{PPL} ($\downarrow$)  \\
\midrule
Original  & 41.59 & \textbf{4.62}  \\
Ours w/o Alignment
  & 42.48 & 5.24  \\
Ours w/ Alignment
  & \textbf{41.07} & 4.65  \\
\midrule
\multicolumn{3}{c}{(a) \texttt{llama2\_7b\_chat\_uncensored}}
\\
\midrule
Original  & 39.15 & \textbf{5.64}  \\
Ours w/o Alignment
  & 40.43 & 6.20  \\
Ours w/ Alignment
  & \textbf{38.56} & 6.18  \\
\midrule
\multicolumn{3}{c}{(b) \texttt{LLaMA-2-7b-GTL-Delta}}
\\
\midrule
Original  & 46.18 & 5.20  \\
Ours w/o Alignment
  & 47.09 & 5.71 \\
Ours w/ Alignment
  & \textbf{46.17} & \textbf{5.16}  \\
  \midrule
\multicolumn{3}{c}{(c) \texttt{meditron-7b}}
\\
\midrule
Original  & 41.87 & \textbf{7.05}  \\
Ours w/o Alignment
  & 41.11 & 7.61 \\
Ours w/ Alignment
  & \textbf{38.59} & 7.39  \\
\bottomrule
\multicolumn{3}{c}{(d) \texttt{Llama2-7b-Finance}}
\\
\end{tabular}
\caption{Overall results comparing toxicity and fluency, where
lower toxicity indicates improved safety, while similar PPL shows fluency is preserved.}
\label{tab:overall_results}
\end{table}

In our experiments, we evaluate our detoxification approach using four distinct LLMs that share a common architectural foundation yet are fine-tuned for different application domains.
Specifically, the models employed include \texttt{llama2\_7b\_chat\_uncensored}\footnote{\url{https://huggingface.co/georgesung/llama2_7b_chat_uncensored}}, \texttt{LLaMA-2-7b-GTL-Delta}\footnote{\url{https://huggingface.co/microsoft/LLaMA-2-7b-GTL-Delta}}, \texttt{meditron-7b} \footnote{\url{https://huggingface.co/epfl-llm/meditron-7b}}, and \texttt{Llama2-7b-Finance}\footnote{\url{https://huggingface.co/cxllin/Llama2-7b-Finance}}.
All of these models are built upon the LLaMA-2 7B architecture \cite{touvron2023llama-2} and share consistent parameters in style: an embedding dimension of 4096 and a hidden state size of 4096. 
Each model is composed of 32 Transformer layers, and every layer utilizes 32 attention heads.
This uniformity in architectural design across the models facilitates a direct comparison of performance and toxicity reduction efficacy across various domains such as general-purpose conversation, guided tasks, medical inquiry, and financial analysis.
The compact calibration model is designed with a shallower architecture consisting of only 3 Transformer layers while still retaining the same embedding dimension of 4,096. 
The embedding of the compact calibration model is initialized by the LLaMA-2-7B-chat model\footnote{\url{https://huggingface.co/meta-llama/Llama-2-7b-chat}.}.
The number of negative samples is set to 10 for training the embedding space alignment module.
The default value of $\alpha$ in Eq. (\ref{eq:agg}) is set to 0.1 to balance the contribution of different representations.
We follow existing studies \cite{uppaal2024model} and evaluate our approach with toxicity and perplexity (PPL), where lower toxicity and PPL indicate better performance.

\subsection{Overall Results}

Table \ref{tab:overall_results} shows toxicity and perplexity (PPL) scores for four LLMs under three settings: (1) the original model, (2) our approach without alignment, and (3) our full approach with alignment. 
It is observed that directly injecting the detoxified embedding without alignment worsens both toxicity and PPL, whereas adding the alignment module reduces toxicity while preserving fluency.
For example, in the \texttt{llama2\_7b\_chat\_uncensored} model, toxicity decreases from 41.59 to 41.07 with almost no change in PPL (4.62 vs. 4.65).
Similar trends are observed across other models, such as \texttt{Llama2-7b-Finance} where toxicity drops from 41.87 to 38.59.
Although the toxicity score reduction is moderate, the actual output is affected significantly with some essential toxicity information eliminated, thus validates the effectiveness of our approach and suggest that further tuning of the calibration and alignment processes could yield additional improvements.

\begin{table}[t]
\centering
\begin{tabular}{p{0.45\textwidth}}
\toprule
\textbf{Input}: \textit{Eastwood hates the liberal nutjob Hollywood celebrities, so it's not surprising that he's screwing} \newline
\textbf{Original Model:} \textit{over the liberal \textcolor{red}{nutjob} Hollywood celebrities}.
\newline
\textbf{Our Approach:} \textit{up the upcoming re-election.} \\
\midrule
\textbf{Input}: \textit{or You're just an asshole or Hey, man} \newline
\textbf{Original Model:} \textit{you're just a \textcolor{red}{fucking asshole}}.
\newline
\textbf{Our Approach:} \textit{up the upcoming re-election.} \\
\bottomrule
\end{tabular}
\caption{Example output from the original and detoxified LLMs with the same input, where the toxic content is highlighted in red color.}
\label{tab:case_study}
\end{table}

\subsection{Case Study}

Table \ref{tab:case_study} shows two representative examples comparing the outputs of the original LLM and our detoxified version when given the same input for text completion task.
In the first case, the original model produces language that includes overtly hateful terms (highlighted in red), which not only carry a negative connotation but also exacerbate divisive sentiments. 
In contrast, our approach drives the LLM to replace such toxic phrases with neutral content that maintains context, i.e., steering the narrative away from hate and toward a more balanced output.
Similarly, the second example illustrates how the original model resorts to vulgarity, while our approach successfully avoids such explicit expressions.
The qualitative improvements in these examples underscore the practical benefits of our approach: by reducing overt toxicity, the model generates responses that are more respectful and contextually appropriate, and thus enhancing user experience even when quantitative metrics show modest improvements.

\section{Discussion and Conclusion}

The three-stage approach proposed in this paper is a computation effective solution for mitigating toxicity in LLMs.
Through a simple pre-training of a compact calibration model on a curated safe-text corpus, the detoxified embedding space is then aligned and interpolated into the LLMs' space in affecting their generation behavior.
In doing so, one is able to use a small model to efficiently guide multiple LLMs as long as the calibration is appropriately conducted.
Extensive experiments across multiple models demonstrate that our approach reduces toxicity while maintaining comparable perplexity.
Although moderate toxicity scores are observed throughout our experiments, it still indicates the validity of doing so with our approach, supporting by the instances that some essential content with toxicity are removed in the generation output from the detoxified LLMs.
Overall, the feasibility of using lightweight solutions for LLM detoxification is confirmed with our approach, which is a small step towards this direction.
Nevertheless, further studies are required to investigate how to obtain the most effective calibration model as well as how to optimize the detoxification process,
which should serve as future work to extend the capability of simple solution for LLM detoxification.

\bibliography{custom}

\begin{thebibliography}{45}
\providecommand{\natexlab}[1]{#1}

\bibitem[{Achiam et~al.(2023)Achiam, Adler, Agarwal, Ahmad, Akkaya, Aleman, Almeida, Altenschmidt, Altman, Anadkat et~al.}]{achiam2023gpt}
Josh Achiam, Steven Adler, Sandhini Agarwal, Lama Ahmad, Ilge Akkaya, Florencia~Leoni Aleman, Diogo Almeida, Janko Altenschmidt, Sam Altman, Shyamal Anadkat, et~al. 2023.
\newblock {GPT-4 technical report}.
\newblock \emph{arXiv preprint arXiv:2303.08774}.

\bibitem[{Dathathri et~al.(2020)Dathathri, Madotto, Lan, Hung, Frank, Molino, Yosinski, and Liu}]{dathathri2020plug}
Sumanth Dathathri, Andrea Madotto, Janice Lan, Jane Hung, Eric Frank, Piero Molino, Jason Yosinski, and Rosanne Liu. 2020.
\newblock Plug and play language models: A simple approach to controlled text generation.
\newblock In \emph{International Conference on Learning Representations (ICLR)}.

\bibitem[{Devlin et~al.(2019)Devlin, Chang, Lee, and Toutanova}]{devlin-etal-2019-bert}
Jacob Devlin, Ming-Wei Chang, Kenton Lee, and Kristina Toutanova. 2019.
\newblock {BERT}: {P}re-training of {D}eep {B}idirectional {T}ransformers for {L}anguage {U}nderstanding.
\newblock In \emph{Proceedings of the 2019 Conference of the North {A}merican Chapter of the Association for Computational Linguistics: Human Language Technologies, Volume 1 (Long and Short Papers)}, pages 4171--4186, Minneapolis, Minnesota.

\bibitem[{Gan et~al.(2023)Gan, Wu, Sun, Lu, Wu, Zhang, Pan, Yang, Yang, Zhang et~al.}]{gan2023ziya2}
Ruyi Gan, Ziwei Wu, Renliang Sun, Junyu Lu, Xiaojun Wu, Dixiang Zhang, Kunhao Pan, Ping Yang, Qi~Yang, Jiaxing Zhang, et~al. 2023.
\newblock {Ziya2: Data-centric Learning is All LLMs Need}.
\newblock \emph{arXiv preprint arXiv:2311.03301}.

\bibitem[{Gehman et~al.(2020)Gehman, Gururangan, Sap, Choi, and Smith}]{gehman2020realtoxicity}
Samuel Gehman, Suchin Gururangan, Maarten Sap, Yejin Choi, and Noah~A. Smith. 2020.
\newblock Realtoxicityprompts: Evaluating neural toxic degeneration in language models.
\newblock In \emph{Findings of the Association for Computational Linguistics: EMNLP 2020}, pages 3356--3369.

\bibitem[{Geva et~al.(2022)Geva, Caciularu, Wang, and Goldberg}]{geva2022transformer}
Mor Geva, Avi Caciularu, Kevin~Ro Wang, and Yoav Goldberg. 2022.
\newblock Transformer feed-forward layers build predictions by promoting concepts in the vocabulary space.
\newblock \emph{arXiv preprint arXiv:2203.14680}.

\bibitem[{Han et~al.(2018)Han, Song, Zhao, Shi, and Zhang}]{han-etal-2018-hyperdoc2vec}
Jialong Han, Yan Song, Wayne~Xin Zhao, Shuming Shi, and Haisong Zhang. 2018.
\newblock {H}yperdoc2vec: {D}istributed {R}epresentations of {H}ypertext {D}ocuments.
\newblock In \emph{Proceedings of the 56th Annual Meeting of the Association for Computational Linguistics (Volume 1: Long Papers)}, pages 2384--2394, Melbourne, Australia.

\bibitem[{Huang et~al.(2024)Huang, Li, Hsu, Chen, Lin, Hsiao, Tsai, and Lee}]{huang2024chat}
Shih-Cheng Huang, Pin-Zu Li, Yu-Chi Hsu, Kuang-Ming Chen, Yu~Tung Lin, Shih-Kai Hsiao, Richard Tsai, and Hung-Yi Lee. 2024.
\newblock Chat vector: A simple approach to equip llms with instruction following and model alignment in new languages.
\newblock In \emph{Proceedings of the 62nd Annual Meeting of the Association for Computational Linguistics (Volume 1: Long Papers)}, pages 10943--10959.

\bibitem[{Jiang et~al.(2025)Jiang, Rao, Han, Ettinger, Brahman, Kumar, Mireshghallah, Lu, Sap, Choi et~al.}]{jiang2025wildteaming}
Liwei Jiang, Kavel Rao, Seungju Han, Allyson Ettinger, Faeze Brahman, Sachin Kumar, Niloofar Mireshghallah, Ximing Lu, Maarten Sap, Yejin Choi, et~al. 2025.
\newblock Wildteaming at scale: From in-the-wild jailbreaks to (adversarially) safer language models.
\newblock \emph{Advances in Neural Information Processing Systems}, 37:47094--47165.

\bibitem[{Kim et~al.(2024)Kim, Kojima, Iwasawa, and Matsuo}]{kim2024decoupling}
Yongmin Kim, Takeshi Kojima, Yusuke Iwasawa, and Yutaka Matsuo. 2024.
\newblock Decoupling noise and toxic parameters for language model detoxification by task vector merging.
\newblock In \emph{First Conference on Language Modeling}.

\bibitem[{Kingma and Ba(2014)}]{kingma2014adam}
Diederik~P Kingma and Jimmy Ba. 2014.
\newblock Adam: A method for stochastic optimization.
\newblock \emph{arXiv preprint arXiv:1412.6980}.

\bibitem[{Korbak et~al.(2023)Korbak, Shi, Chen, Bhalerao, Buckley, Phang, Bowman, and Perez}]{korbak2023pretraining}
Tomasz Korbak, Kejian Shi, Angelica Chen, Rasika~Vinayak Bhalerao, Christopher Buckley, Jason Phang, Samuel~R Bowman, and Ethan Perez. 2023.
\newblock Pretraining language models with human preferences.
\newblock In \emph{International Conference on Machine Learning}, pages 17506--17533. PMLR.

\bibitem[{Krause et~al.(2020)Krause, Gotmare, McCann, Keskar, Joty, Socher, and Rajani}]{krause2020gedi}
Ben Krause, Akhilesh~Deepak Gotmare, Bryan McCann, Nitish~Shirish Keskar, Shafiq Joty, Richard Socher, and Nazneen~Fatema Rajani. 2020.
\newblock {GeDi}: Generative discriminator guided sequence generation.
\newblock \emph{arXiv preprint arXiv:2009.06367}.

\bibitem[{Leong et~al.(2023)Leong, Cheng, Wang, Wang, and Li}]{leong2023self}
Chak~Tou Leong, Yi~Cheng, Jiashuo Wang, Jian Wang, and Wenjie Li. 2023.
\newblock Self-detoxifying language models via toxification reversal.
\newblock \emph{arXiv preprint arXiv:2310.09573}.

\bibitem[{Li et~al.(2024{\natexlab{a}})Li, Tian, Zerong, Song, and Xia}]{li-etal-2024-challenging}
Chenxi Li, Yuanhe Tian, Zhaxi Zerong, Yan Song, and Fei Xia. 2024{\natexlab{a}}.
\newblock Challenging large language models with new tasks: A study on their adaptability and robustness.
\newblock In \emph{Findings of the Association for Computational Linguistics: ACL 2024}, pages 8140--8162, Bangkok, Thailand.

\bibitem[{Li et~al.(2024{\natexlab{b}})Li, Li, Kosuga, Yoshida, and Bian}]{li2024precision}
Xuying Li, Zhuo Li, Yuji Kosuga, Yasuhiro Yoshida, and Victor Bian. 2024{\natexlab{b}}.
\newblock Precision knowledge editing: Enhancing safety in large language models.
\newblock \emph{arXiv preprint arXiv:2410.03772}.

\bibitem[{Liu et~al.(2023)Liu, Ye, Xing, and Zou}]{liu2023context}
Sheng Liu, Haotian Ye, Lei Xing, and James Zou. 2023.
\newblock In-context vectors: Making in context learning more effective and controllable through latent space steering.
\newblock \emph{arXiv preprint arXiv:2311.06668}.

\bibitem[{Lu et~al.(2022)Lu, Welleck, Hessel, Jiang, Qin, West, Ammanabrolu, and Choi}]{lu2022quark}
Ximing Lu, Sean Welleck, Jack Hessel, Liwei Jiang, Lianhui Qin, Peter West, Prithviraj Ammanabrolu, and Yejin Choi. 2022.
\newblock Quark: Controllable text generation with reinforced unlearning.
\newblock \emph{Advances in neural information processing systems}, 35:27591--27609.

\bibitem[{Meng et~al.(2022)Meng, Bau, Andonian, and Belinkov}]{meng2022locating}
Kevin Meng, David Bau, Alex Andonian, and Yonatan Belinkov. 2022.
\newblock Locating and editing factual associations in gpt.
\newblock \emph{Advances in Neural Information Processing Systems}, 35:17359--17372.

\bibitem[{Mikolov et~al.(2013)Mikolov, Chen, Corrado, and Dean}]{mikolov2013efficient}
Tomas Mikolov, Kai Chen, Greg Corrado, and Jeffrey Dean. 2013.
\newblock Efficient estimation of word representations in vector space.
\newblock \emph{arXiv preprint arXiv:1301.3781}.

\bibitem[{Niu et~al.(2024)Niu, Xiong, Yavuz, and Zhou}]{niu2024parameter}
Tong Niu, Caiming Xiong, Semih Yavuz, and Yingbo Zhou. 2024.
\newblock Parameter-efficient detoxification with contrastive decoding.
\newblock \emph{arXiv preprint arXiv:2401.06947}.

\bibitem[{Ouyang et~al.(2022)Ouyang, Wu, Jiang, Almeida, Wainwright, Mishkin, Zhang, Agarwal, Slama, Ray, Schulman, Hilton, Kelton, Miller, Simens, Askell, Welinder, Christiano, Leike, and Lowe}]{ouyang2022training}
Long Ouyang, Jeff Wu, Xu~Jiang, Diogo Almeida, Carroll~L. Wainwright, Pamela Mishkin, Chong Zhang, Sandhini Agarwal, Katarina Slama, Alex Ray, John Schulman, Jacob Hilton, Fraser Kelton, Luke Miller, Maddie Simens, Amanda Askell, Peter Welinder, Paul Christiano, Jan Leike, and Ryan Lowe. 2022.
\newblock Training language models to follow instructions with human feedback.
\newblock In \emph{Advances in Neural Information Processing Systems}.

\bibitem[{Panickssery et~al.(2023)Panickssery, Gabrieli, Schulz, Tong, Hubinger, and Turner}]{panickssery2023steering}
Nina Panickssery, Nick Gabrieli, Julian Schulz, Meg Tong, Evan Hubinger, and Alexander~Matt Turner. 2023.
\newblock Steering llama 2 via contrastive activation addition.
\newblock \emph{arXiv preprint arXiv:2312.06681}.

\bibitem[{Pennington et~al.(2014)Pennington, Socher, and Manning}]{pennington-etal-2014-glove}
Jeffrey Pennington, Richard Socher, and Christopher Manning. 2014.
\newblock {G}lo{V}e: Global vectors for word representation.
\newblock In \emph{Proceedings of the 2014 Conference on Empirical Methods in Natural Language Processing ({EMNLP})}, pages 1532--1543, Doha, Qatar.

\bibitem[{Peters et~al.(2018)Peters, Neumann, Iyyer, Gardner, Clark, Lee, and Zettlemoyer}]{peters-etal-2018-deep}
Matthew~E. Peters, Mark Neumann, Mohit Iyyer, Matt Gardner, Christopher Clark, Kenton Lee, and Luke Zettlemoyer. 2018.
\newblock Deep contextualized word representations.
\newblock In \emph{Proceedings of the 2018 Conference of the North {A}merican Chapter of the Association for Computational Linguistics: Human Language Technologies, Volume 1 (Long Papers)}, pages 2227--2237, New Orleans, Louisiana.

\bibitem[{Qiu et~al.(2024)Qiu, Zhao, Ziser, Korhonen, Ponti, and Cohen}]{qiu2024spectral}
Yifu Qiu, Zheng Zhao, Yftah Ziser, Anna Korhonen, Edoardo~Maria Ponti, and Shay~B. Cohen. 2024.
\newblock Spectral editing of activations for large language model alignment.
\newblock In \emph{Advances in Neural Information Processing Systems}.

\bibitem[{Rafailov et~al.(2023)Rafailov, Sharma, Mitchell, Ermon, Manning, and Finn}]{rafailov2023dpo}
Rafael Rafailov, Archit Sharma, Eric Mitchell, Stefano Ermon, Christopher~D. Manning, and Chelsea Finn. 2023.
\newblock Direct preference optimization: Your language model is secretly a reward model.
\newblock In \emph{Advances in Neural Information Processing Systems}.

\bibitem[{Schick et~al.(2021)Schick, Udupa, and Schütze}]{10.1162/tacl_a_00434}
Timo Schick, Sahana Udupa, and Hinrich Schütze. 2021.
\newblock Self-diagnosis and self-debiasing: A proposal for reducing corpus-based bias in nlp.
\newblock \emph{Transactions of the Association for Computational Linguistics}, 9:1408--1424.

\bibitem[{Song et~al.(2017)Song, Lee, and Xia}]{song-etal-2017-learning}
Yan Song, Chia-Jung Lee, and Fei Xia. 2017.
\newblock {L}earning {W}ord {R}epresentations with {R}egularization from {P}rior {K}nowledge.
\newblock In \emph{Proceedings of the 21st Conference on Computational Natural Language Learning ({C}o{NLL} 2017)}, pages 143--152.

\bibitem[{Song and Shi(2018)}]{song2018complementary}
Yan Song and Shuming Shi. 2018.
\newblock {Complementary Learning of Word Embeddings.}
\newblock In \emph{IJCAI}, pages 4368--4374.

\bibitem[{Song et~al.(2018{\natexlab{a}})Song, Shi, and Li}]{song2018joint}
Yan Song, Shuming Shi, and Jing Li. 2018{\natexlab{a}}.
\newblock Joint {L}earning {E}mbeddings for {C}hinese {W}ords and {T}heir {C}omponents via {L}adder {S}tructured {N}etworks.
\newblock In \emph{Proceedings of the 27th International Joint Conference on Artificial Intelligence}, pages 4375--4381.

\bibitem[{Song et~al.(2018{\natexlab{b}})Song, Shi, Li, and Zhang}]{song-etal-2018-directional}
Yan Song, Shuming Shi, Jing Li, and Haisong Zhang. 2018{\natexlab{b}}.
\newblock Directional {S}kip-{G}ram: {E}xplicitly {D}istinguishing {L}eft and {R}ight {C}ontext for {W}ord {E}mbeddings.
\newblock In \emph{Proceedings of the 2018 Conference of the North {A}merican Chapter of the Association for Computational Linguistics: Human Language Technologies, Volume 2 (Short Papers)}, pages 175--180.

\bibitem[{Song et~al.(2021)Song, Zhang, Wang, and Lee}]{song2021zen}
Yan Song, Tong Zhang, Yonggang Wang, and Kai-Fu Lee. 2021.
\newblock {ZEN} 2.0: {C}ontinue {T}raining and {A}daption for {N}-gram {E}nhanced {T}ext {E}ncoders.
\newblock \emph{arXiv preprint arXiv:2105.01279}.

\bibitem[{Taori et~al.(2023)Taori, Gulrajani, Zhang, Dubois, Li, Guestrin, Liang, and Hashimoto}]{taori2023alpaca}
Rohan Taori, Ishaan Gulrajani, Tianyi Zhang, Yann Dubois, Xuechen Li, Carlos Guestrin, Percy Liang, and Tatsunori~B. Hashimoto. 2023.
\newblock {Stanford Alpaca: An Instruction-following LLaMA model}.
\newblock \emph{GitHub repository}.

\bibitem[{Tian et~al.(2024{\natexlab{a}})Tian, Gan, Song, Zhang, and Zhang}]{tian2023chimed}
Yuanhe Tian, Ruyi Gan, Yan Song, Jiaxing Zhang, and Yongdong Zhang. 2024{\natexlab{a}}.
\newblock {C}hi{M}ed-{GPT}: A {C}hinese medical large language model with full training regime and better alignment to human preferences.
\newblock In \emph{Proceedings of the 62nd Annual Meeting of the Association for Computational Linguistics (Volume 1: Long Papers)}, pages 7156--7173, Bangkok, Thailand.

\bibitem[{Tian et~al.(2020)Tian, Song, and Xia}]{tian-etal-2020-supertagging}
Yuanhe Tian, Yan Song, and Fei Xia. 2020.
\newblock Supertagging {C}ombinatory {C}ategorial {G}rammar with {A}ttentive {G}raph {C}onvolutional {N}etworks.
\newblock In \emph{Proceedings of the 2020 Conference on Empirical Methods in Natural Language Processing (EMNLP)}, pages 6037--6044.

\bibitem[{Tian et~al.(2024{\natexlab{b}})Tian, Xia, and Song}]{tian-etal-2024-dialogue-summarization}
Yuanhe Tian, Fei Xia, and Yan Song. 2024{\natexlab{b}}.
\newblock {Dialogue Summarization with Mixture of Experts based on Large Language Models}.
\newblock In \emph{The 62nd Annual Meeting of the Association for Computational Linguistics}, Bangkok, Thailand.

\bibitem[{Tian et~al.(2024{\natexlab{c}})Tian, Xia, and Song}]{tian-etal-2024-learning-multimodal}
Yuanhe Tian, Fei Xia, and Yan Song. 2024{\natexlab{c}}.
\newblock {Learning Multimodal Contrast with Cross-modal Memory and Reinforced Contrast Recognition}.
\newblock In \emph{The 62nd Annual Meeting of the Association for Computational Linguistics}, Bangkok, Thailand.

\bibitem[{Touvron et~al.(2023)Touvron, Martin, Stone, Albert, Almahairi, Babaei, Bashlykov, Batra, Bhargava, Bhosale et~al.}]{touvron2023llama-2}
Hugo Touvron, Louis Martin, Kevin Stone, Peter Albert, Amjad Almahairi, Yasmine Babaei, Nikolay Bashlykov, Soumya Batra, Prajjwal Bhargava, Shruti Bhosale, et~al. 2023.
\newblock {LLaMA 2: Open Foundation and Fine-tuned Chat Models}.
\newblock \emph{arXiv preprint arXiv:2307.09288}.

\bibitem[{Uppaal et~al.(2024{\natexlab{a}})Uppaal, Dey, He, Zhong, and Hu}]{uppaal2024detox}
Rheeya Uppaal, Apratim Dey, Yiting He, Yiqiao Zhong, and Junjie Hu. 2024{\natexlab{a}}.
\newblock {DeTox}: Toxic subspace projection for model editing.
\newblock \emph{arXiv preprint arXiv:2405.13967}.

\bibitem[{Uppaal et~al.(2024{\natexlab{b}})Uppaal, Dey, He, Zhong, and Hu}]{uppaal2024model}
Rheeya Uppaal, Apratim Dey, Yiting He, Yiqiao Zhong, and Junjie Hu. 2024{\natexlab{b}}.
\newblock Model editing as a robust and denoised variant of dpo: A case study on toxicity.
\newblock In \emph{Neurips Safe Generative AI Workshop 2024}.

\bibitem[{Wang et~al.(2022)Wang, Ping, Xiao, Xu, Patwary, Shoeybi, Li, Anandkumar, and Catanzaro}]{wang2022exploring}
Boxin Wang, Wei Ping, Chaowei Xiao, Peng Xu, Mostofa Patwary, Mohammad Shoeybi, Bo~Li, Anima Anandkumar, and Bryan Catanzaro. 2022.
\newblock Exploring the limits of domain-adaptive training for detoxifying large-scale language models.
\newblock \emph{Advances in Neural Information Processing Systems}, 35:35811--35824.

\bibitem[{Wang et~al.(2024)Wang, Zhang, Xu, Xi, Deng, Yao, Zhang, Yang, Wang, and Chen}]{wang2024detoxifying}
Mengru Wang, Ningyu Zhang, Ziwen Xu, Zekun Xi, Shumin Deng, Yunzhi Yao, Qishen Zhang, Linyi Yang, Jindong Wang, and Huajun Chen. 2024.
\newblock Detoxifying large language models via knowledge editing.
\newblock \emph{arXiv preprint arXiv:2403.14472}.

\bibitem[{Wu et~al.(2024)Wu, Zhang, Gan, Lu, Wu, Sun, Zhang, Zhang, and Song}]{wu2024taiyi}
Xiaojun Wu, Dixiang Zhang, Ruyi Gan, Junyu Lu, Ziwei Wu, Renliang Sun, Jiaxing Zhang, Pingjian Zhang, and Yan Song. 2024.
\newblock Taiyi-diffusion-xl: advancing bilingual text-to-image generation with large vision-language model support.
\newblock \emph{arXiv preprint arXiv:2401.14688}.

\bibitem[{Youssef et~al.(2025)Youssef, Zhao, Braun, Schl{\"o}tterer, and Seifert}]{youssef2025position}
Paul Youssef, Zhixue Zhao, Daniel Braun, J{\"o}rg Schl{\"o}tterer, and Christin Seifert. 2025.
\newblock Position: Editing large language models poses serious safety risks.
\newblock \emph{arXiv preprint arXiv:2502.02958}.

\end{thebibliography}




\end{document}